\documentclass[conference]{IEEEtran}
\IEEEoverridecommandlockouts
\usepackage{cite}
\usepackage{amsmath,amssymb,amsfonts}
\usepackage{algorithmic}
\usepackage{graphicx}
\usepackage{textcomp}

\usepackage{graphics} 
\usepackage[]{algorithm2e}
\usepackage{epsfig} 
\usepackage{mathptmx} 
\usepackage{amsmath} 
\usepackage{balance}
\usepackage{multirow}
\usepackage{comment}

\usepackage{makecell}

\def\BibTeX{{\rm B\kern-.05em{\sc i\kern-.025em b}\kern-.08em
    T\kern-.1667em\lower.7ex\hbox{E}\kern-.125emX}}
\begin{document}

\title{Moody Learners - Explaining Competitive Behaviour of Reinforcement Learning Agents}

 \author{\IEEEauthorblockN{Pablo Barros\IEEEauthorrefmark{1},
 Ana Tanevska\IEEEauthorrefmark{1},
 Francisco Cruz\IEEEauthorrefmark{2}\IEEEauthorrefmark{3},
 Alessandra Sciutti\IEEEauthorrefmark{1} }
 \IEEEauthorblockA{\IEEEauthorrefmark{1}Cognitive Architecture for Collaborative Technologies Unit (CONTACT),\\
 Italian Institute of Technology (IIT), Genova, Italy\\ 
 Email: \{pablo.alvesdebarros, ana.tanevska, alessandra.sciutti\}@iit.it}
 \IEEEauthorblockA{\IEEEauthorrefmark{2} Escuela de Ingenier\'ia, Universidad Central de Chile, Santiago, Chile}
 \IEEEauthorblockA{\IEEEauthorrefmark{3} School of Information Technology, Deakin University, Geelong, Australia\\
 Email: francisco.cruz@deakin.edu.au}
 }


\maketitle

\begin{abstract}

Designing the decision-making processes of artificial agents that are involved in competitive interactions is a challenging task. In a competitive scenario, the agent does not only have a dynamic environment but also is directly affected by the opponents' actions. 
Observing the Q-values of the agent is usually a way of explaining its behavior, however, it does not show the temporal-relation between the selected actions. We address this problem by proposing the Moody framework that creates an intrinsic representation for each agent based on the Pleasure/Arousal model. We evaluate our model by performing a series of experiments using the competitive multiplayer Chef's Hat card game and discuss how by observing the intrinsic state generated by our model allows us to obtain a holistic representation of the competitive dynamics within the game.

\end{abstract}


\section{Introduction}

The application of reinforcement learning (RL) models has been on a rise following the onset of deep reinforcement learning \cite{mnih2013playing}. The ability of RL models to solve complex problems, represented mostly by the association of high-dimensional states and a large number of discrete or continuous actions, has recently led to the development of expert systems for guiding autonomous cars \cite{sallab2017deep, isele2018navigating}, predicting the stock exchange impact \cite{ponomarev2019using, meng2019reinforcement}, and coordinating a swarm of robots to protect the environment \cite{haksar2018distributed, yu2018reinforcement}, to name a few examples.

As their popularity grows, the need for providing stable and trustworthy solutions for real-world problems also increases. When deployed in production environments, these RL models, although designed to achieve impressive performance, are not easily understandable, and thus need experts to address the problems that may arise. The robustness of the RL solutions is thus limited by the general inability of explaining easily how these models learn and derive their state-action mapping \cite{cruz2019memory}. This has led to the fast development of eXplainable AI (XAI)  \cite{gunning2017explainable} as a complementary field for deep reinforcement learning. 

The community around XAI has been investigating how to approach the understanding of reinforcement learning when applied to different problems. Problems that involve the processing of human-based data are somehow the easiest ones to address, as demonstrated by explaining recommendation systems \cite{wang2018reinforcement}, applications on the medical domain \cite{holzinger2017we}, and robotics applications based on known behavior \cite{anjomshoae2019explainable}. However, it is much more difficult to explain the agents' behavior in scenarios where the environment cannot be easily modeled \cite{goebel2018explainable}, or the solutions are unknown a priori \cite{sheh2017did}. In this regard, recent applications derive human-level explanation based on the agent's own knowledge of the situation \cite{cruz2020explainable}. In particular, transforming the selected Q-values for each action into a confidence metric, using a re-shaping function based on the logarithmic transformation \cite{cruz2020explainable}, improved the understanding of a robot trying to solve a grid-based navigation task. The confidence metric measures how a specific action contributes to the robot reaching its goal however it does not carry any temporal correlation between the actions selected by the agent during the navigation. This means that the metric cannot be used to explain the behavior of the agent within the entire simulation, but only for each action. 

This approach functions quite well when the end state of the task is clearly defined, and it is possible to measure the distance between the current state and the end state. However, it does not deal well with competitive scenarios where a set of agents have to learn decisions that a) maximize their goal, and b) minimize their adversaries' goals. Besides considering the dynamic of the scenarios, they usually have to deal with the interactions between the agents themselves. Some of the most common applications for competitive reinforcement learning involve the design and implementation of learning agents in simulations that simplify several aspects of the real world such as the implementation of grid-based autonomous vehicles \cite{fridman2018deeptraffic}, life-simulation/resources gathering \cite{xu2018hierarchical}, and multi-player games \cite{mckenzie2017competitive}. In these cases, explaining the agents' behavior based on the simplified environment is somehow possible, but they would not scale for real-world applications.

In this regard, we propose a novel methodology to explain the behavior of artificial agents in real-world modeling a competitive scenario. We deploy our experiments onto the  multiplayer Chef's Hat card game \cite{barros2020chef} as it offers a dynamic interaction between the players and simulates directly the real-world counterpart game. In the scenarios' baseline, four agents play against each other and their performance is measured directly based on how many games they win \cite{barrosICPR2020}. It is not possible, however, without an extensive manual observation of their action-selection pattern to explain their winning behavior while the game happens.

To identify the impact of each selected action, we propose the \emph{Moody framework} which uses the agents' own evaluation of the game to provide a temporal reference for the impact of selected actions and translate it into an pleasure/arousal representation  \cite{costa1996mood} of the agent's mood. Our method builds on the introspective transformation from the Q-values \cite{cruz2020explainable}, and introduces a Growing-When-Required (GWR) network to transform confidence readings into the pleasure/arousal scale, and to create strong inter-action correlation, which directly maps the action-selection-mood triplet in our competitive card game scenario and carries a temporal momentum that we use to explain the agents' performance while the game happens.

We also investigate the \emph{Moody framework}  as a tool for explaining the agents' behavior under a competitive perspective. The intrinsic mood readings are directly formulated by an agent assessing its own actions. In a competitive scenario, the actions of the agent's  opponents impact directly the agents' own mood. Thus, being informed on how each agent perceives their opponent's actions can give us a much richer explanation of the performance of the agent while the game happens. 

In this paper, we formalize the entire \emph{Moody framework}, and aim on addressing two main research questions: 1) How the mood readings provide the understanding of the agents' behavior on the Chef's Hat game; 2) How close is the representation of the mood estimated by an agent about its opponents from the real mood of each of these opponents. To answer these questions we perform a series of experiments where different agents play a series of games using the Chef's Hat simulation environment.  We explain how the agents' own assessment of the opponents' actions impacts the performance explanation. Also, to better understand the impact of the \emph{Moody framework} on the XAI community, we discuss the effect of our self-evaluation in competitive scenarios and how they can provide a closed-world representation of the entire game.

\section{Game Mechanics and Agents Implementation}
\subsection{The Chef's Hat Card Game}

The game setup which we used as an environment for our artificial agents is the Chef's Hat card game \cite{barros2020food}. Chef's Hat as a game provides a controllable action-perception cycle, where each player can only perform a restricted set of actions. This in turn allows each player to behave as organically as possible, and allows us to directly measure the impact of each action within the naturally-controllable real-world scenario. 

Chef's Hat is a 4-player round-based card game, where each person has a restaurant-context role (Chef, Sous-Chef, Waiter or Dishwasher), that is updated after each game based on the order of finishing the previous match. At the beginning of the game the players get the full hand of cards (17 cards per player) dealt to them, and taking turns they need to dispose of their cards as quickly as possible. The cards represent ingredients for pizzas, and each round consists of the players making pizzas by discarding cards in a certain manner. The details of the pizza-making rules and the role hierarchy are explained fully in the games' formal description \cite{barros2020food}. In order to better explain our model, we detail the flow of one full game in Algorithm \ref{alg:ChefsHat}. 

\begin{algorithm}
 Shuffle the deck; \\
Deal an equal amount of cards per player; \\
  Exchange roles; \\
  Exchange cards; \\
   \If{special action is evoked}
  {
    Do special action;
  }
  
$FirstPlayer\gets Has golden 11$\\
 $FirstPlayer$ discard cards. \\
 
 \While{not end of the game}{
  \For{ each player}{

      \eIf{player can, and want, to discard}{
       discard cards\;
       }{
        pass\;
      }
      \If{All players passed}
      {
        Make the pizza;
        $FirstPlayer\gets Last player to discard$\\
      }
        \If{All players finished}
      {
        End of game.
      }
   }
 }
 \caption{The Game-flow of the Chef's Hat card game.}
 \label{alg:ChefsHat}
\end{algorithm}

\subsection{The Chef's Hat Simulator Implementation}
\begin{figure}[h]
    \centering
  \includegraphics[width=0.65\columnwidth]{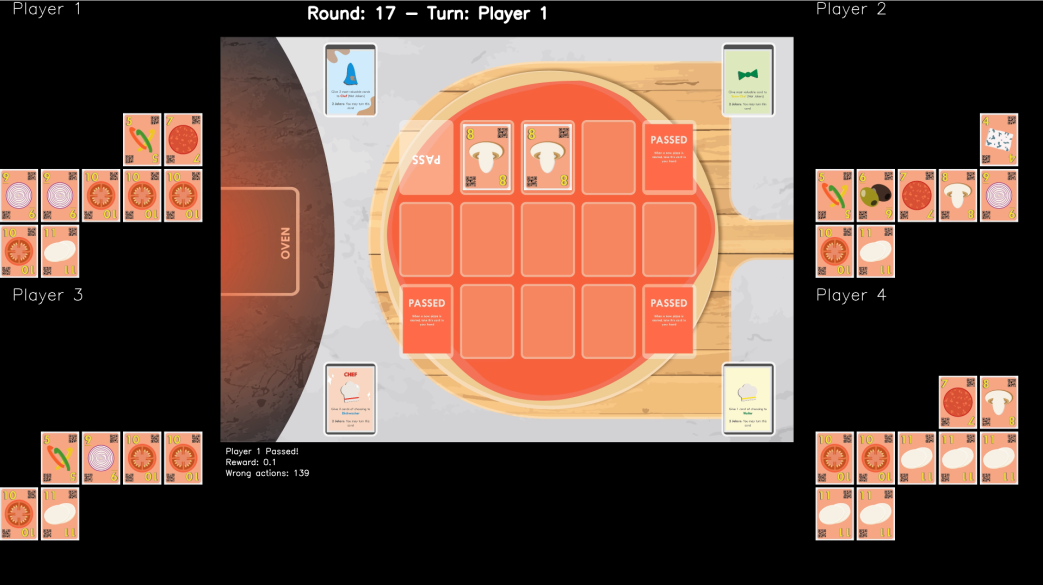}
  \caption{Chef's Hat rendered simulation environment.}
\label{fig:gameExample}
\end{figure}

We implemented our scenario using the OpenAI-based simulation environment of Chef's Hat \cite{barros2020chef}. The environment simulates all of the game mechanics, and thus provides a 1:1 simulation of the real-world game, and provides an easy implementation of different agents to play the game, as shown in Figure \ref{fig:gameExample}. 

The current game state for each player is represented as an aggregation of the cards of the player and the current cards in the playing field for a total of 28 values. Each player can choose among 200 different actions to perform in each state, each of them representing a unique combination of cards to be discarded, or a pass action.

In order for the simulated players to be able to select their actions in an intelligent manner, we needed to train artificial agents to play Chef's Hat. For this we employ two different reinforcement learning algorithms based on Q-learning, which allow our agents to apply a temporal-difference calculation when updating the policy network and thus maximize the state transitions that will lead to the optimal reward.

\subsection{The Learning Agents}

The main characteristic of the \emph{Moody Framework} is to provide a self-assessment of the agents' performance based on the agent's own judgments. To better evaluate that, in our experiments, we implement three different types of agents, two learning ones based on Deep Q-Learning - DQL \cite{van2016deep} and on on Proximal Policy Optimization - PPO \cite{schulman2017proximal}, and a dummy one which only select random actions. To guarantee that a taken action by one agent is valid, we use the proposed Chef's Hat action selection greedy policy which limits each agents' final action selection by applying a mask on the final Q-values composed of only the currently allowed actions, given a certain state.

As discussed by Barros et al. \cite{barrosICPR2020}, the learning agents learn different strategies when trained to play the Chef's Hat game. While the DQL agent learns a set of restricted actions, which are mostly deployed by the end of the match, the PPO agent usually has higher confidence in winning the game when using a set of actions at the beginning of the match.

\begin{figure}[h]
    \centering
    \includegraphics[width=0.65\columnwidth]{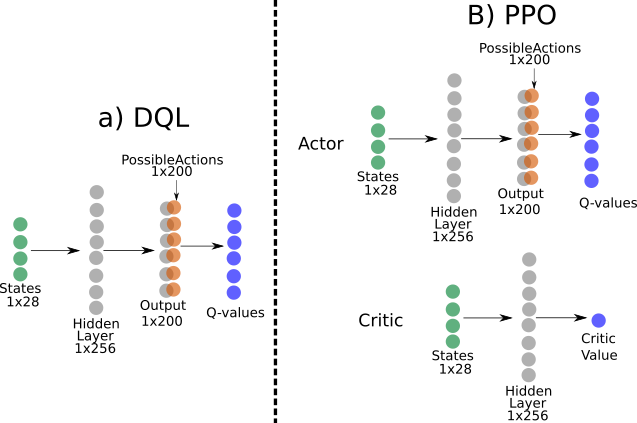}
    \caption{The detailed implementation of the DQL and the PPO agents.}
    \label{fig:models}
\end{figure}

In this research, we are interested in establishing an explanation on how each of these agents performs while playing Chef's Hat, so we do not focus or detail the agents' learning processes. To guarantee that each agent has its own strategies when playing the game, we follow the baseline protocols established by the \emph{vs Everyone} self-play routine \cite{barrosICPR2020}. Figure \ref{fig:models} illustrates our implemented agents.

\subsection{The Principles of Q-Learning}

Both of our agents implement a Q-learning routine. The general Q-learning algorithm learns to maximize the probability of choosing an action that leads to maximum reward. For that, it calculates a Q-value (quality value) for each action given a state and updates the policy, in our case represented by a neural network, to maximize the expected reward. Using a temporal-difference calculation, it can take into consideration a sequence of steps that leads to the final state, represented by finishing all cards in your hand. The maximal reward is given once the player is the first one to reach the final state.

The typical Q-learning algorithm represents a function $Q$:

\begin{equation}
Q:SxA \rightarrow \mathbb{R}
\end{equation}
where $S$ is the state, in our case represented by the 28 values composed by the 17 cards at hand and the 11 cards at the board. The actions, $A$, are expressed using the 200 discrete values for all the possible actions.

To update the Q-values, the algorithm uses the following update function:

\begin{equation}
    Q'(s_t, a_t)= Q(s,a) + \alpha  \times \left (  TD \right )
\end{equation}
where $t$ is the current step, $\alpha$ is a pre-defined learning rate and $TD$ is the temporal-difference function, calculated as:

\begin{equation}
TD=r_t \times \gamma \times maxQ \left ( s_{t+1}, a_t \right ) - maxQ \left ( s_t, a_t \right )
\end{equation}
where $r_t$ is the obtained reward for the state ($s_t$) and action ($a_t$) association, $\gamma$ represents the discount factor, a modulator that estimates the importance of the future rewards, and $maxQ\left (s_{t+1}, a_t \right )$ is the estimation of Q-value for the next state.

\section{A \emph{Moody Framework} }

To evaluate the performance of the agents when playing Chef's Hat is not a simple task. At first, state-based evaluations could give us a straight answer to performance measures, but they do need external evaluation and they would be completely biased on the evaluators' knowledge about the game. Counting the number of cards a player has at hand at a specific moment can indicate a precise performance value, but it can vary drastically depending on the players' own strategy. The same goes for measuring number of discarded cards in each turn. 

Having a self-evaluation of the agents' own performance, thus, could solve this problem. As the action-selection strategy of the agent is made based on its own knowledge, the agent will be able to explain its own actions based on its own judgment. 


One of the simplest ways of explaining the action-selection impact of a player is a Q-value-based observation. The Q-value represents the impact that each of the 200 actions will have, given a certain state, to reach the maximal goal. For a given state, the agent calculates which action would give it the highest probability of reaching the end state. The Q-values reading, however, does not allow us to establish how closely the agent is of winning the game by choosing that action, and thus, does not give us a better understanding of the agents' performance, but only of which action was the most appropriate given that specific state.

\subsection{Calculating Confidence}

To address the problem of calculating the impact of the selected action towards reaching the final goal, the introspection-based transformation of the Q-values \cite{cruz2020explainable} was proposed. This approach focuses on scaling the selected Q-value towards the final goal using a logarithm transformation which computes the probability of success, that we use as a confidence ($C$), that this specific action will lead the agent to reach the final state:

\begin{equation}
C = \left ( \frac{1}{2} \times log_{10}  \frac{Q(s,a)}{R^{T}} \right )
\end{equation}

where $Q(s,a)$ is the Q-value of a selected action, and $R^{T}$ is the maximum reward achieved by the agent at that specific time step. To normalize the probability between the interval $[0,1]$, the probability is saturated and every value smaller than 0 is set to 0, and every value above 1 is set to 1.

The probability is calculated based on how well that Q-value scales towards the maximum reward. In the Chef's Hat game, the agent only reaches the final reward if it wins a game. Also, for each action which is not the final reward, the agent gets a -0.01 reward in order to encourage the agent to prefer shorter paths to the final reward \cite{barrosICPR2020}. This impacts directly on the final reward calculation, as for each new turn the agent is playing, the final reward changes. To take this into consideration, we update the final reward per turn following:

\begin{equation}
R^{T} = 1 - \left( T\times0.01 \right )
\end{equation}
where $T$ represents the current turn on that specific game.
 
 \subsection{A Moody neural network}

 The observation of the Q-values on the Chef's Hat scenario already allowed for an understanding of how different learning mechanisms impacted on the final performance of each agent during a series of games \cite{barros2020chef}. Given the Chef's Hat specific greedy policy, however, each action that an agent selects is limited by the possible actions given that specific state. There is a strong limitation on the actions an agent can select, and thus, it impacts directly on the Q-values obtained for that specific state.

\begin{figure}
\centering
\includegraphics[width=1\columnwidth]{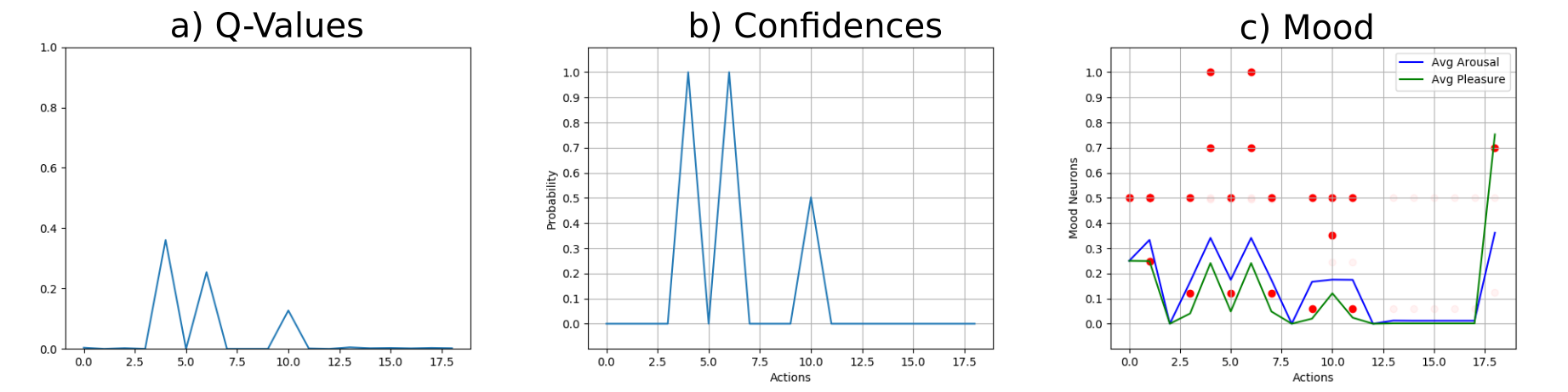}
\caption{Examples of the Q-values (a), confidence (b), and mood (c) readings of the same agent on the same game. In the mood readings (c), each of the dots represent a neuron of the network after an action was performed. As less transparent the neuron, as higher its habituation counter, meaning it was used recently to represent a perceived confidence.}
\label{fig:readingsExample}
\end{figure}

 In a certain state, only a handful of possible actions are allowed and the agent has to establish which of those actions provide the best way to reach the maximum goal. This translates to the agent having a very small Q-value per action, as the allowed actions do not always comprise on the best ones. We illustrate this behavior on the readings of Figure \ref{fig:readingsExample} (a). This is especially impacted by the competitive characteristic of the game. The agent's next action will not be impacted only by its own previous action, but by the combined impact of the actions of its opponents. 
 
 Evaluating them alone in the Chef's Hat competitive scenario, thus, give us a very poor reading of the game context, as it only represents how that specific action has the highest probability, among all the allowed actions, to reach the final state. It does not carry any information about how this action was impacted by the previous ones. The same can be said by the confidence values, as they are a direct transformation of the Q-values. Although they can give us a clearer picture of how that specific action is contributing to the agent reaching the maximal goal, it will be extremely sensitive to the fast changes of the competitive dynamics of the game, as illustrated in the example plotted in Figure \ref{fig:readingsExample} (b).
 
 To address that and in order to obtain a grounded representation of the agent's own internal state, we propose the implementation of a Growing-When-Required Network (GWR) \cite{Marsland2002} to generate prototypical representations of the impact of the measured confidences. The GWR was used recently to address personalization on emotion expression perception \cite{barros2017self, barros2019personalized}, and in several continuous learning tasks \cite{Parisi2017, Parisi2018}. It is a self-organizing network that creates neurons that represents a series of input data and it can be trained online to approximate to the perceived stimuli.
 
Before feeding it to the GWR, we transform the calculated confidences into a pleasure/arousal (PA) scale \cite{costa1996mood}, which represents a perceived experience two dimensions: pleasing/unpleasing and excited/calm. The PA model has been used as a standard way of representing intrinsic states in virtual agents \cite{pena2011representing, zhang2016modeling, shvo2019towards}, and will allow an easy understanding our \emph{Moody framework} representations, and improve its applicability in different scenarios.

Each neuron of the GWR has a weight vector ${w}_j$ that represents a prototype of the perceived information. In our scenario, we feed the GWR with perceived PA values. A newly perceived PA value will be associated with a best-matching unit (BMU) $b$, which is calculated by minimizing the distances between the perceived PA value and all the neurons on the GWR. Given a set of $N$ neurons, $b$ concerning the input ${x}\in\mathbb{R}^n$ is computed as:

\begin{equation}
\begin{aligned}
b = \arg\min_{j\in N} \left(\Vert {x} - {w}_j  \Vert^2 \right).
\end{aligned}
\end{equation}

When the BMU is found, new connections are created between the BMU and the second-BMU. Every neuron that is connected to the BMU is its topological neighbour. Each neuron has an aging mechanism, represented by an habituation counter $h_i \in [0,1]$, which represents how close this neuron was to the BMU, and thus, how important it is for representing the current input.

The habituation rule is given by:

\begin{equation}
\begin{aligned}
\Delta h_i=\tau_i \cdot \kappa \cdot (1-h_i)-\tau_i.
\end{aligned}
\end{equation}
where $\kappa$ and $\tau_i$ are constants that control the decreasing behavior of the habituation counter \cite{Marsland2002}. To establish whether a neuron is habituated, its habituation counter $h_i$ must be smaller than a given habituation threshold $t_h$.

The network is initialized with two neurons and, at each learning iteration, it inserts a new neuron whenever the activity of the network when a confidence $c$ is calculated, of a habituated neuron, is smaller than a given threshold $t_a$, i.e., a new neuron is created if $a(c)<t_a$ and $h_c<t_h$. The activity of the network is given by:

\begin{equation}
\begin{aligned}
a(c) = \exp(-\left(\Vert {x} - {w}_j  \Vert^2 \right)
\end{aligned}
\end{equation}

For each perceived PA value, the network updates, in an online manner, the state of the neurons by updating the BMUs and their neighbors, or adding new neurons. Neurons that are old, with a smaller habituation counter than a certain threshold, are removed from the network. That guarantees a dynamic behavior that represents the PA values as soon as they are perceived, and although the model has not a direct temporal processing mechanism, such as recurrent connections, we are able to maintain a temporal relation on the perceived PA values by adding neurons that represent PA values never perceived before and removing neurons which represent PA values that were not perceived anymore.

\subsection{Assessing my own Mood}

The goal of the \emph{Moody framework}  is to provide a self-assessment, coming from the agent itself, about its own actions. The GWR will generate a temporal momentum on the confidences that will represent the agent's performance based on its own judgment. To achieve it, we calculate the Q-values, and subsequently the confidences, of all the allowed actions over a given state. We then sum all these confidences and represent the agents' confidence that it can reach the final state, given the cards it has at hand and the board at that moment. The last step is to convert the confidence value into a PA scale. We then train the network with the PA values for each action taken by the agent.

Different from the typical clustering applications, the GWR on the \emph{Moody framework} is a stateful model. We obtain the mood reading by averaging all the weights of all the neurons of the network at that specific state. As the neurons change over the game, giving the agents' own estimation, the mood readings will increase or decrease depending on how that agent perceives its own actions. Most importantly, the mood reading will carry the temporal relations between the estimations. This is observed by the example plotted in Figure \ref{fig:readingsExample} (c), where we also observe the temporal behavior of the neurons been inserted in the network to map the PA values. We clearly see when a neuron with higher PA value starts to get more active, and thus, has a smaller habituation. 

The Chef's Hat card game is made to be played in a series of games. The roles assignments are given based on the finishing position of the last game, and they change the general games' winning probabilities. When playing with completely random agents, each agent has a chance of 25\% of winning the game. However, if this agent won the previous game and it receives the Chef role in the current game, the chances of winning the game increase to 35\% \cite{barros2020food}.
To represent the differences between the self-assessed confidence over an action and the environment given feedback when finishing a game, we calculate the PA values as follows:

\begin{equation}
\begin{aligned}
P = \left\{
\begin{matrix}
 C \times 0.5& if & action \\ 
1  & if & victory \\ 
 0& if & notVictory 
\end{matrix}
\right.
\end{aligned}
\end{equation}

\begin{equation}
\begin{aligned}
A = \left\{
\begin{matrix}
 0.5 - ((C-0.5)/0.25) & if & action & and & C < 0.5 \\ 
0.5 + ((C-0.5)/0.25) & if &  action & and &C\geq 0.5 \\ 
C 1 & if &  victory \\ 
C 0 & if &  victory \\ 
\end{matrix}
\right.
\end{aligned}
\end{equation}

 The pleasure (P) scale represents how pleased the agent is given a certain event, and for every action the agent takes, we correlate it directly to the confidence, reducing its impact by 50\%. The arousal scale impacts the agent's own perception of how excited it is when performing an action. Again, to modulate the weight of the agents' own confidence in its actions, every action will give the agent value of 0.5 (neutral excitement) and will be modulated by the agent's own assessment. As closer as the confidence is of 1, as higher is the excitement modulation that the agent perceives. The opposite happens when confidence is less than 0.5. 
Obtaining a victory at the end of the game should give the agent a clear pleasure and arousal signal, and thus, have to be the strongest signal that the agent receives. 

The proposed different modulation for pleasure and arousal presented here are to be taken as one perspective on how to interpret the impact of the confidence. Different modulations can be designed and derived, but it will not be the focus of this paper. The use of the proposed modulation will not impact our final considerations.






\subsection{Assessing my Opponent's Mood}
 
The same way an agent can assess its mood, it can also assess its opponents' mood. This is important to give us a complete information about the agent's perception of the entire competitive game, taking into consideration how it assesses its actions, and how it assess the others' actions. To achieve the others' assessment, we propose the use of an estimated confidence (E-Confidence), which is calculated using a partial estimation of the opponents' hands.
 
To calculate its self-confidence, the agent uses the full game state, composed of the cards it has at hand and the board. When observing an opponent, the agent does not have access to the opponent's hand, only to the cards on the board. The agent can, however, compose an estimated hand based on the amount of cards the opponent already discarded and the current cards the opponent played on the field. The estimated hand is composed of 17 cards, and each of them is calculated as follows:
 
\begin{equation}
\begin{aligned}
card = \left\{
\begin{matrix}
0 & if & noCard\\
d & if & discarded \\
r & if & card At Hand
\end{matrix}
\right.
\end{aligned}
\end{equation}

where $d$ represents the discarded  cards by the opponent on that round, if any, and $r$ is a randomly chosen card, between the face value interval $[1,11]$. To normalize the estimation, we create 100 different combinations of ($e-C$). Knowing which action the opponent took, the agent will calculate the estimated confidence value for that specific action for the 100 combinations. This will give the agent an approximated assessment of the opponent's action, based on its judgment. 

\begin{figure*}
    \centering
    \includegraphics[width=1.5\columnwidth]{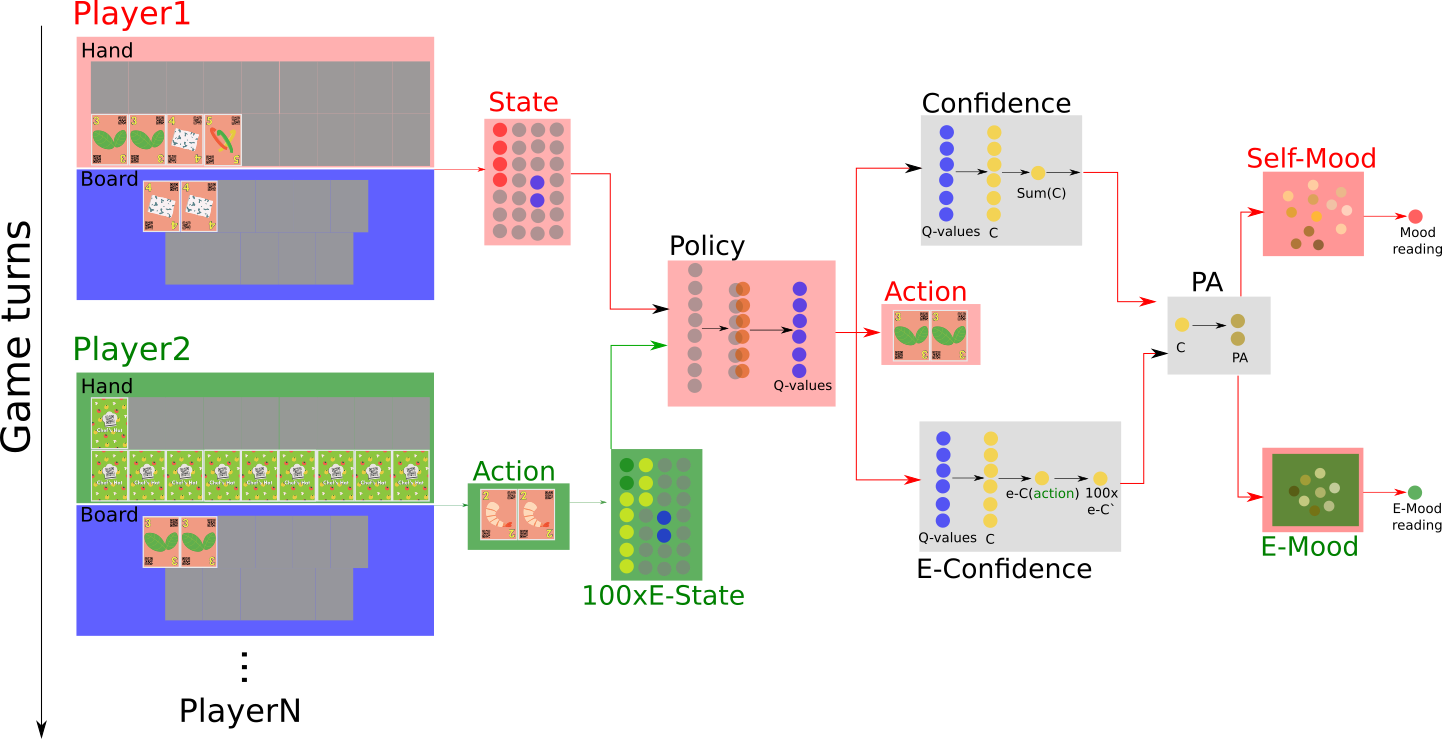}
    \caption{Illustration of how Player 1 (boxes with the red shade) obtains its own mood reading and the estimated mood reading of Player 2 (boxes with the green shade). Player 1 calculates its own states (represented by the remaining cards at hand (in red), the cards on the board (in blue), and the empty card slots (in gray)), estimates its  Q-values and confidences, updates its mood and selects an action, so the board is updated. Player 1 estimates the state of Player 2 by using the cards in the board (blue dots), the discarded cards (green dots), and by estimating 100 combinations for each of the missing cards (light-green dots). It calculates the Q-values for each of the e-states, and subsequently the e-confidences which are used to update the e-mood for Player 2.}
    \label{fig:moodyFramework}
\end{figure*}

For each of the agent opponents, we create a single GWR to represent an estimated mood. For each opponent's action, we create a specific $e-{PA}$, following the same procedure demonstrated in Eq. 9 and 10, and train one specific GWR for that opponent. The entire process provides each agent with its own mood readings and the estimated mood readings of each of its opponents. Figure \ref{fig:moodyFramework} illustrates the entire processing flow of the \emph{Moody framework} , illustrating how the Player 1 obtains its mood reading and the e-mood reading for one opponent.

\section{Experiments}

We perform two rounds of experiments in order to address our two main research questions: 1) How the mood readings provide the understanding of the agents' behavior on the Chef's Hat game; 2) How close is the representation of the mood estimated by an agent about its opponents from the real mood of each of these opponents.


\subsection{Mood vs Confidence}
 
 Our first experiment aims to demonstrate the capabilities of the \emph{Moody framework} to describe the performance of an agent when making actions on the Chef's Hat game. To do so, we set up one game and put a DQL-based, a PPO-based and 2 dummy agents to play against each other. We collect the confidence and mood values for the DQL and PPO agents and exhibit how they describe each other's behavior during the game. We also collect and exhibit the estimated confidence and estimated mood that each of these two agents calculated about each other. To illustrate further the capabilities of the mood, we also exhibit the same measures for 10 games played in a row.

\subsection{Self vs Estimated}

To evaluate how well an agent can estimate the performance of another agent within the same game, we run 100 games where 2 DQL-based agents play against 2 PPO-based agents. We calculate the correlation between the mood and confidences and the estimated mood and confidences obtained by each of the agents about themselves.  Using two agents with the same learning algorithm playing against each other will allow us to evaluate also the impact of the learned strategies on how to play the game on the estimated values.

\section{Results and Discussions}

\subsection{Mood vs Confidence}

\begin{figure}
    \centering
    \includegraphics[width=1\linewidth]{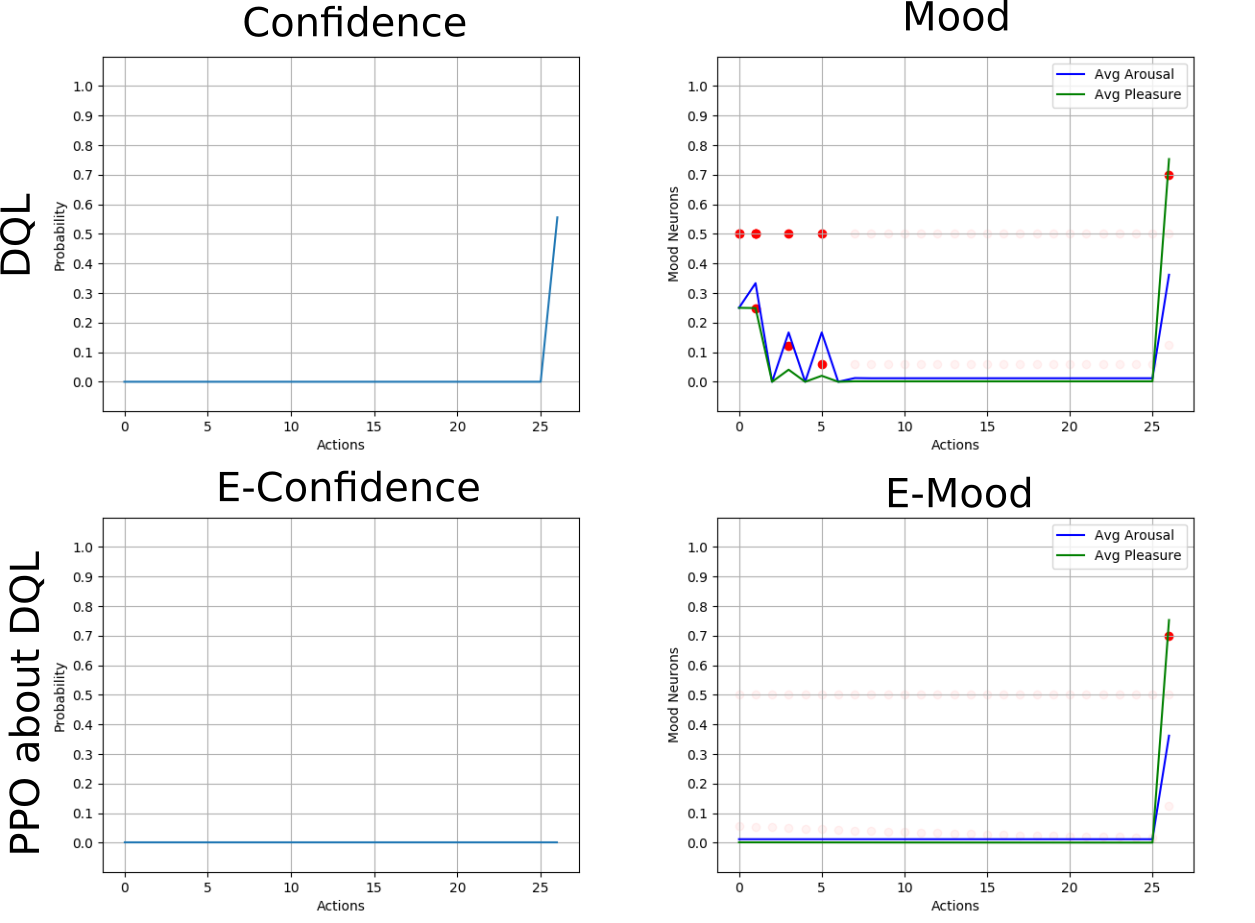}
    \caption{Confidence and mood of the DQL-based agent, and e-confidence and e-mood estimated by PPO-based agent for 1 game won by the DQL-based agent.}
    \label{fig:dqlM1}
\end{figure}

The winner of the single match was the DQL agent. It won after performing 26 actions. The confidence and mood of the DQL-based agent are exhibited in Figure \ref{fig:dqlM1}, together with the estimation of the DQL confidences and mood obtained by the PPO-based agent. It is easy to see the inherent behavior enforced by the DQL algorithm on showing high confidence on actions that happen by the end of the game, as explained in the analysis made by Barros et al. \cite{barros2020chef}. Most importantly, we observe that, in this game, the last action performed by DQL was assessed as having strong confidence. The mood readings map this behavior clearly, presenting a higher pleasure and arousal reading by the end of the match, which is enhanced by this agent winning the match. 

The estimation made by PPO-based about the DQL-based agent, at first, seems different from the estimations obtained by DQL-based agent itself. However, once understanding how the PPO agent learns to play the game, as explained in detail by Barros et al. \cite{barros2020chef}, the estimations seem to make more sense. The PPO-based agent develops during training a strategy that enforces it to have higher confidence in its actions at the beginning of the game when compared to the end of the game. That means the PPO-based agent evaluates the actions from the DQL-based agent using its understanding of what good actions are. And thus, presents a very particular version of DQL-based agents' confidence, disregarding the last action as having a higher impact.

\begin{figure}
    \centering
    \includegraphics[width=1\linewidth]{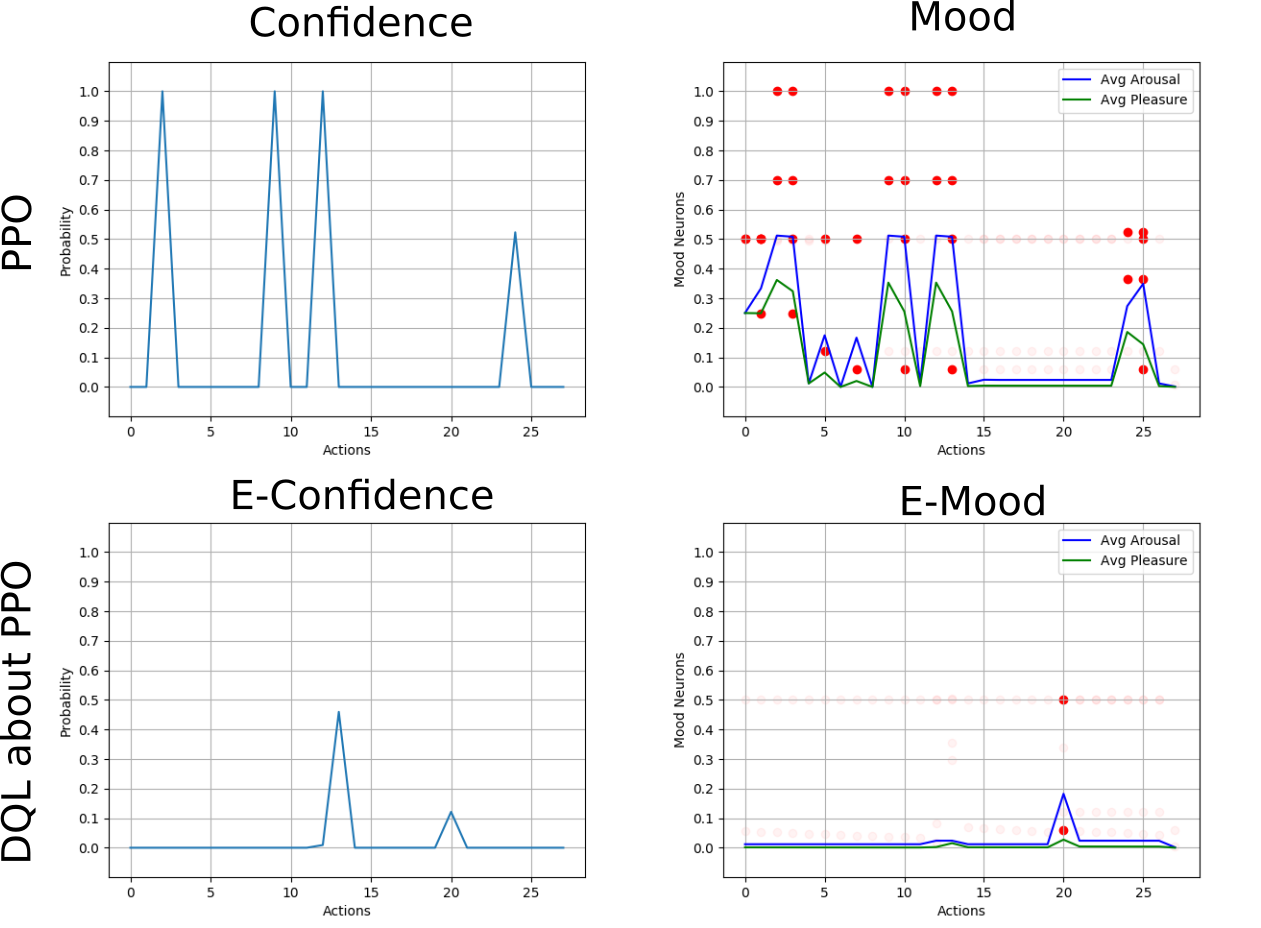}
    \caption{Confidence and mood of the PPO-based agent, and e-confidence and e-mood estimated by the DQL-based agent for 1 game won by the DQL-based agent.}
    \label{fig:ppoM1}
\end{figure}

Doing the same observation on the plots about the PPO-based agent, illustrated in Figure \ref{fig:ppoM1}, we find the same behavior. The DQL-based agent estimates the PPO-based agent's confidence based on its interpretation of what a good action is, and thus, creates its version of the PPO-based agent performance. In both cases, however, it is much easier to understand the behavior of each of these agents during the game by observing the mood readings. Even when observing the estimated readings, we can clearly identify how well these agents are doing in the game, and in which point it happened a turnover that allowed the DQL-based agent to win the game, at around action 25.

\begin{figure}
    \centering
    \includegraphics[width=1\linewidth]{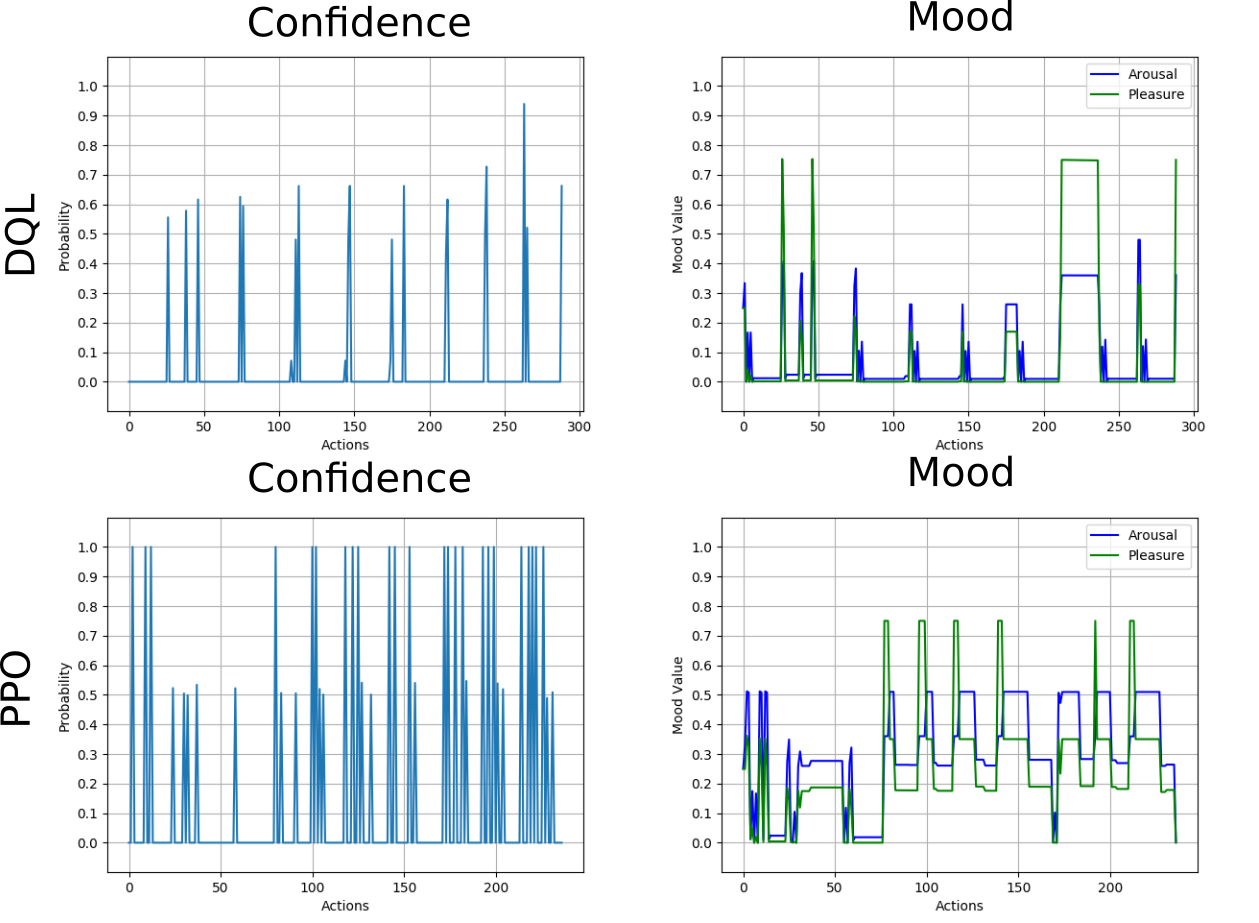}
    \caption{Confidence and mood of the DQL-based agent and the PPO-based for 10 games, 3 of them won by the DQL-based agent and 7 won by the PPO-based agent.}
    \label{fig:bothM10}
\end{figure}

The advantages of using the mood readings to represent an agents' performance instead of the confidence values become much clearer when we report the results of playing 10 games in a row in Figure \ref{fig:bothM10}. By plotting the confidence readings over all these matches, the lack of temporal momentum in-between the actions and games is clearly evident. When observing the mood, however, we can clearly identify when the DQL-based agent is performing well, and win the matches mostly in the beginning of the games, while the PPO-based agent has a winning streak by the end of the 10 games.

\subsection{Self vs Estimated}

In order to give each agent its own closed-world observation of the entire Chef's Hat game, 
it is extremely important that the estimated readings are somehow trustworthy. Of course, given that each agent learns to derive playing strategies differently \cite{barros2020chef}, it is clear that their assessment of each other's actions are different. Calculating the correlation of the mood readings and confidence values after running 100 games with two pairs of incarnations of agents that follow the same strategy allow us to validate the agents' estimations. Table \ref{tab:results2} reports all the calculated correlations.

\begin{table}
\centering
\caption{Correlations between estimated confidence and mood readings (averaged between pleasure and arousal) and self-assesed confidence and mood for 100 games played by two DQL-based agents vs two PPO-based agents.}
\centering
\begin{tabular}{c | c c c c}
\hline
\multicolumn{5}{c}{Estimated Confidence}\\\hline
&\multicolumn{4}{c}{From}\\
Target & DQL1 & DQL2 & PPO1 & PPO2 \\\hline
DQL1 & 1 & 0.86 & 0.09& 0.09 \\
DQL2 & 0.77 & 1 & 0.12 & 0.11   \\
PPO1 & 0.12 & 0.12 & 1 & 0.86 \\
PPO2 & 0.04 & 0.04 & 0.86 & 1  \\\hline
\multicolumn{5}{c}{Estimated Mood}\\\hline
&\multicolumn{4}{c}{From}\\
Target & DQL1 & DQL2 & PPO1 & PPO2 \\\hline
DQL1 & 1 & 0.84 & 0.12 & 0.16 \\
DQL2 & 0.88 & 1 & 0.24 & 0.13   \\
PPO1 & 0.24 & 0.19 & 1 & 0.89 \\
PPO2 & 0.10 & 0.08 & 0.92 & 1  \\

\end{tabular}
\label{tab:results2}
\end{table}


When the same types of agents are estimating themselves, DQL1 and DQL2, and PPO1 and PPO2, it is clear that there is a strong correlation between the estimations. The mood estimation, however, presents a general higher correlation than confidence. This happens because of the temporal correlation that happens on the GWR training. Due to the noise in the estimations, in particular at the beginning of the match - as there are many more cards to be estimated - the confidence values can show many different readings. The prototype neurons of the GWR smooth these differences by approximating the inputs towards a single BMU representation, which guarantees a much closer estimation.

Although the different learning algorithms have a different assessment of each others' actions, which is clearly observed by the lower correlations between DQL-based and PPO-based agents, they show a general correlation trend. Here, the mood also provided a higher correlation, mostly due to the same reason: the GWR approximates the correlation, and independently of the learning algorithm, it is easier to track the general performance of the agent.

\subsection{Why the Estimations matter?}

\begin{figure}
    \centering
    \includegraphics[width=1\linewidth]{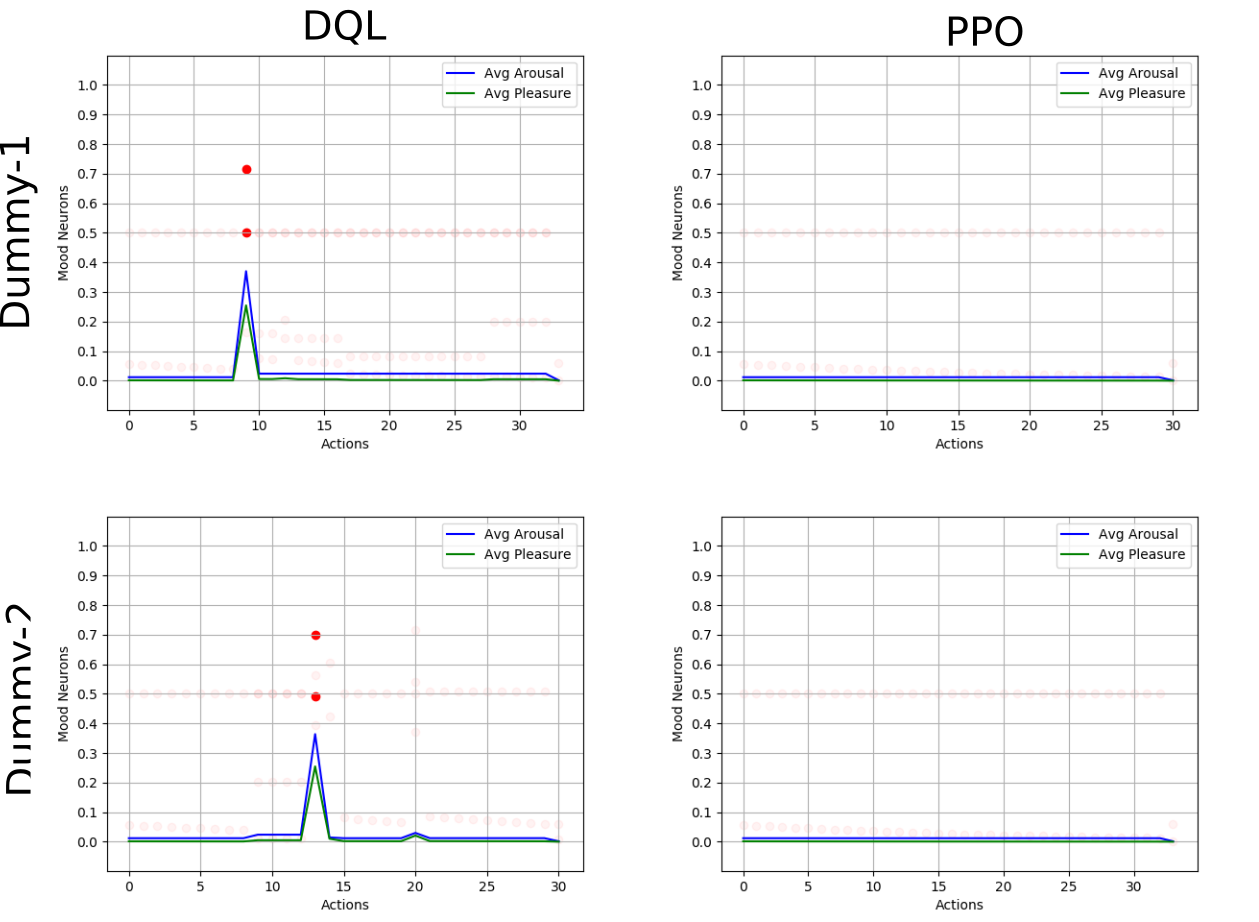}
    \caption{Mood estimation of the DQL and PPO agents when playing one game against two dummy agents.}
    \label{fig:discussion1}
\end{figure}

When analyzing the mood and estimated moods of a single agent, we are able to have an insight into how each of the agents is performing during one game, at an action-selection time. This is clearly demonstrated when visualizing the estimations of DQL-based and PPO-based agents about the dummy agents when playing the single game of the \emph{Self vs Estimated} experiment, illustrated in Figure \ref{fig:discussion1}. It is possible to understand the random behavior, and how it does not contribute at all for winning the game, from both DQL-based and PPO-based agents' perspectives.

Giving to each agent the capability of measuring not only the impact of its own actions but also the estimation of other agents' actions, allows them to have a complete understanding of the game scenario. This gives the agent an important tool that, although not explored in this work, would allow them to interpret any other agents' actions and take this into consideration when performing their own.

\section{Conclusions}

In this paper, we introduced the novel \emph{Moody framework} which is able to explain the behavior of reinforcement learning agents in a competitive multiplayer card game scenario based on the players' assessment of its own performance.

The model is represented in a pleasure and arousal (PA) scale and builds on the introspective confidence \cite{cruz2020explainable} representation of the Q-values selection. It implements a Growing-When-Required (GWR) network to establish a temporal impact between the assessment of the taken actions. Also, the model allows each agent to measure their opponents' actions based on their own assessing, endowing them with a closed-world representation of the entire game. We demonstrate how the \emph{Moody framework} provides a much more enriching explanation of the agents' performance while playing the Chef's Hat card game than the introspection-based confidences. Also, in our experiments, we quantify how well an agent can assess others' actions based on their own judgment.

In general, we see this work as the basis for the implementation of intrinsic-aware agents towards competitive reinforcement learning scenarios. We envision the integration of the mood readings towards an action-selection process in order to modulate the agents' performance based on its own understanding of how well its opponents are performing. Also, we will investigate in the future the integration of intrinsic personality traits in the agents' mood, in order to model unique social relations.

\bibliographystyle{IEEEtran}
\bibliography{bib}

\end{document}